\documentclass[conference]{IEEEtran}
\IEEEoverridecommandlockouts
\usepackage{caption} 
\usepackage{amsmath}
\usepackage{amssymb} 
\usepackage{booktabs}       % professional-quality tables
\usepackage{multirow}
\usepackage{balance}
\usepackage{graphicx}
\usepackage[style=ieee,mincitenames=1,maxcitenames=2,maxbibnames=10,doi=false,isbn=false,url=false,eprint=false]{biblatex}
\DeclareSourcemap{
  \maps[datatype=bibtex, overwrite]{
    \map{
      \step[fieldset=address, null]
      \step[fieldset=location, null]
    }
  }
}

\begin{document}
\title{Visual SLAM with 3D Gaussian Primitives and Depth Priors Enabling Novel View Synthesis}

\author{
\IEEEauthorblockN{1\textsuperscript{st} Zhongche Qu$^{*}$\thanks{$*$ is the corresponding author.}}
\IEEEauthorblockA{\textit{Computer Science} \\
\textit{Columbia University}\\
New York, USA \\
zq2172@columbia.edu
}

~\\
\and
\IEEEauthorblockN{2\textsuperscript{nd} Zhi Zhang}
\IEEEauthorblockA{\textit{Computer Science} \\
\textit{New York University}\\
New York, USA \\
zz2310@nyu.edu
}
~\\

% \and
% \IEEEauthorblockN{3\textsuperscript{rd} Zihan Ye}
% \IEEEauthorblockA{\textit{Technopreneurship and Innovation} \\
% \textit{Nanyang Technological University}\\
% Singapore, Singapore \\
% yezi0004@e.ntu.edu.sg
% }

\and
\IEEEauthorblockN{3\textsuperscript{rd} Cong Liu}
\IEEEauthorblockA{\textit{Computer Science} \\
\textit{Peng Cheng Laboratory}\\
Shenzhen, China \\
liuc@pcl.ac.cn
}

\and
\IEEEauthorblockN{4\textsuperscript{th} Jianhua Yin}
\IEEEauthorblockA{\textit{Computer Science} \\
\textit{Peng Cheng Laborbatory}\\
Shenzhen, China \\
yinjhhust@163.com
}

% \thanks{$*$ is the corresponding author.}
}

\maketitle

%%%%%%%%%%%%%%%%%%%%%%%%%%%%%%%%%%%%%%%%%%%%%%%%%%%%%%%%%%%%%%%%%%%%%%%%%%%%%%%%
\begin{abstract}
Conventional geometry-based SLAM systems lack dense 3D reconstruction capabilities since their data association usually relies on feature correspondences. Additionally, learning-based SLAM systems often fall short in terms of real-time performance and accuracy. Balancing real-time performance with dense 3D reconstruction capabilities is a challenging problem.
In this paper, we present a real-time RGB-D SLAM system that uses 3D Gaussian Splatting for both 3D scene representation and pose estimation. This method enables efficient real-time rendering and differentiable optimization through CUDA, offering an edge over NeRF. We also enable mesh reconstruction from 3D Gaussians for explicit dense 3D reconstruction.
To estimate accurate camera poses, we utilize a rotation-translation decoupled strategy with inverse optimization. This involves iteratively updating both in several iterations through gradient-based optimization. This process includes differentiably rendering RGB, depth, and silhouette maps and updating the camera parameters to minimize a combined loss of photometric loss, depth geometry loss, and visibility loss, given the existing 3D Gaussian map.
However, 3D Gaussian Splatting (3DGS) struggles to accurately represent surfaces due to the multi-view inconsistency of 3D Gaussians, which can lead to reduced accuracy in both camera pose estimation and scene reconstruction. To address this, we utilize depth priors as additional regularization to enforce geometric constraints, thereby improving the accuracy of both pose estimation and 3D reconstruction. We also provide extensive experimental results on public benchmark datasets to demonstrate the effectiveness of our proposed methods in terms of pose accuracy, geometric accuracy, and rendering performance.
\end{abstract}

\begin{figure}[t!]
    \centering
    \includegraphics[width=0.5\textwidth]{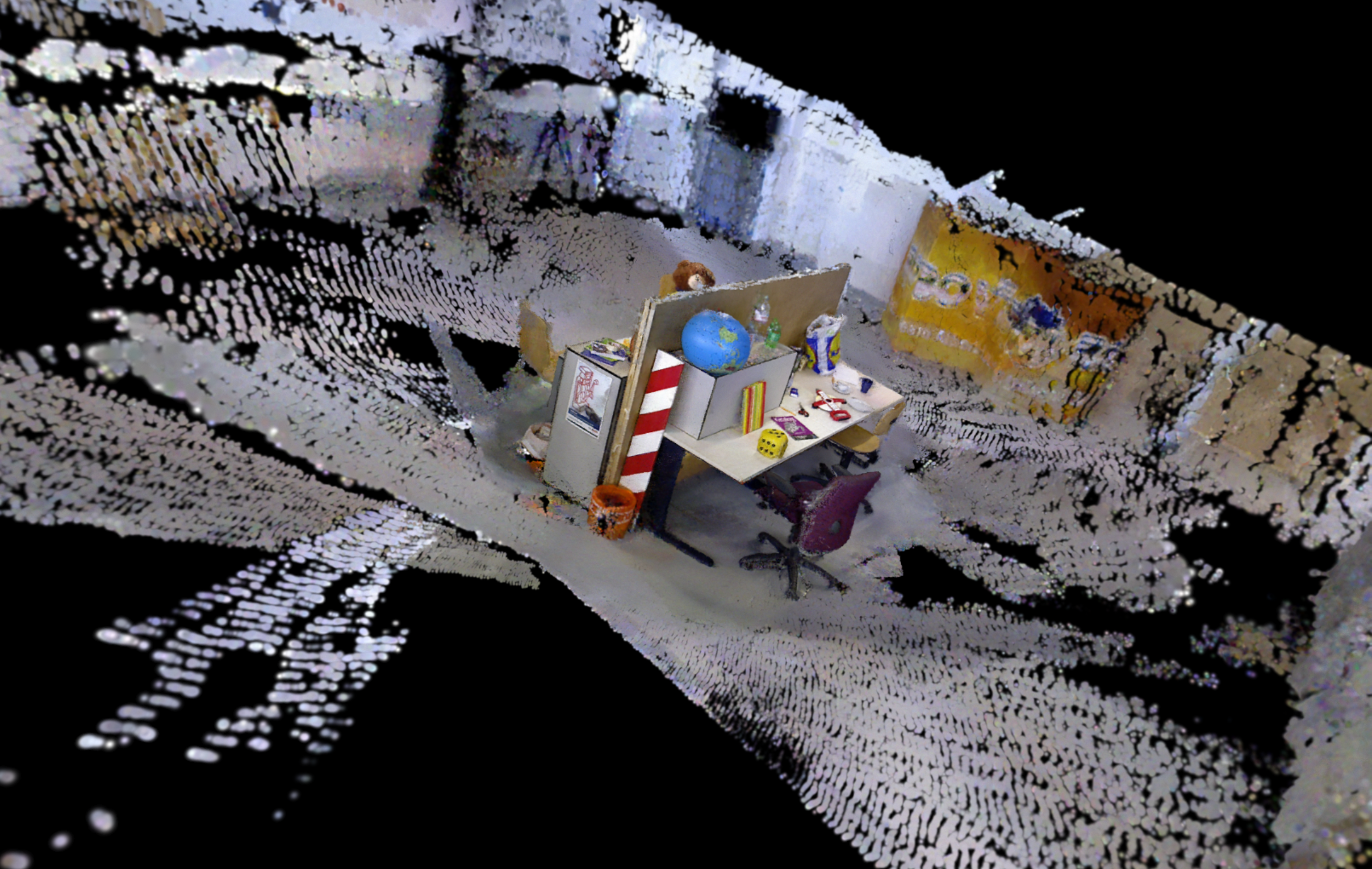}
    \caption{Our final reconstruction results on the Freiburg3 long office sequence.}
    \label{fig:re}
\end{figure}

%%%%%%%%%%%%%%%%%%%%%%%%%%%%%%%%%%%%%%%%%%%%%%%%%%%%%%%%%%%%%%%%%%%%%%%%%%%%%%%%
\section{INTRODUCTION}
% \[{\boldsymbol\gamma}^{\text{b}_\text{k}}_{\text{b}_\text{k+1}}\]
% Gk(x)=αke−12(x−pk)⊤Σ−1k(x−pk)\mathcal{G}_k(\mathbf{x}) = \alpha_k\mathbf{e}^{-\frac{1}{2}(\mathbf{x}-\mathbf{p}_k)^\top\Sigma_k^{-1}(\mathbf{x}-\mathbf{p}_k)}

%\Sigma_k = \mathbf{R}_k\mathbf{s}_k\mathbf{s}_k^\top\mathbf{R}_k^\top 

% \mathbf{c(x)} = \sum_{k=1}^{K}\mathbf{c}_k\alpha_k\mathcal{G}_k^{2D}(\mathbf{x})\prod_{j=1}^{k-1}(1 - \alpha_j\mathcal{G}_j^{2D}(\mathbf{x}))

% \Sigma^\prime = \mathbf{J}\mathbf{W}\Sigma\mathbf{W}^\top\mathbf{J}^\top
Simultaneous Localization and Mapping (SLAM) is a technique for estimating sensor motion and reconstructing the structure of an unknown environment. When SLAM uses cameras, it is specifically referred to as visual SLAM (vSLAM) because it relies solely on visual information. Visual SLAM can be categorized into monocular, stereo, RGB-D, and multi-camera SLAM. Among these, RGB-D SLAM, which provides a depth map as a prior, is particularly useful for many applications such as AR/VR~\cite{edi_wang}, autonomous driving perception~\cite{feng2023robotic, feng2023subsurface, zhang2020manipulator, gao2024decentralized, gao2024adaptive}, 3D reconstruction~\cite{Chen_2023_CVPR, icra23_weihan, ral_jinglun}, novel view synthesis~\cite{Li_2024_CVPR} and navigation~\cite{icia_jinglun,ral_zheng} due to its potential to generate a dense 3D map. Bayesian methods are important in SLAM for providing a framework of handling uncertainty and being able to make predictions from incomplete or noisy data~\cite{zhuang2022defending, zhuang2022robust}. The SLAM problem is often formulated in probabilistic terms, where Bayes' theorem is used to update the knowledge of the robot about its position and the map any time new sensor data becomes available.

% Convential SLAM works
Current research, highlighted by the comprehensive investigations of Ming et al. ~\cite{ming2024benchmarking} in NeRF-based SLAM systems. They found that traditional SLAM faces challenges in matching sensor observations with 3D landmark points, making geometry estimation difficult for unobserved regions and potentially leading to tracking loss. Implicit neural representations, like NeRF, use multi-layer perceptrons (MLPs) to represent the scene. In addition, research fields including Edge computing and Reinforcement learning also integrate well on SLAM. Edge computing ~\cite{li2023predictive, zong2022cocktail} involves processing data near the source of data generation, such as on the robot itself or on a nearby device in the network. Reinforcement learning ~\cite{10406020, li2023red} can be used to develop optimal navigation strategies that allow the robot to efficiently explore unknown areas while avoiding obstacles. By learning from its environment, the robot can make decisions that balance exploration and exploitation. These approaches can complement SLAM by providing quick, localized processing while still allowing for cloud-based analysis if needed.

\begin{figure*}[t]
    \centering
    \includegraphics[width=0.9\textwidth]{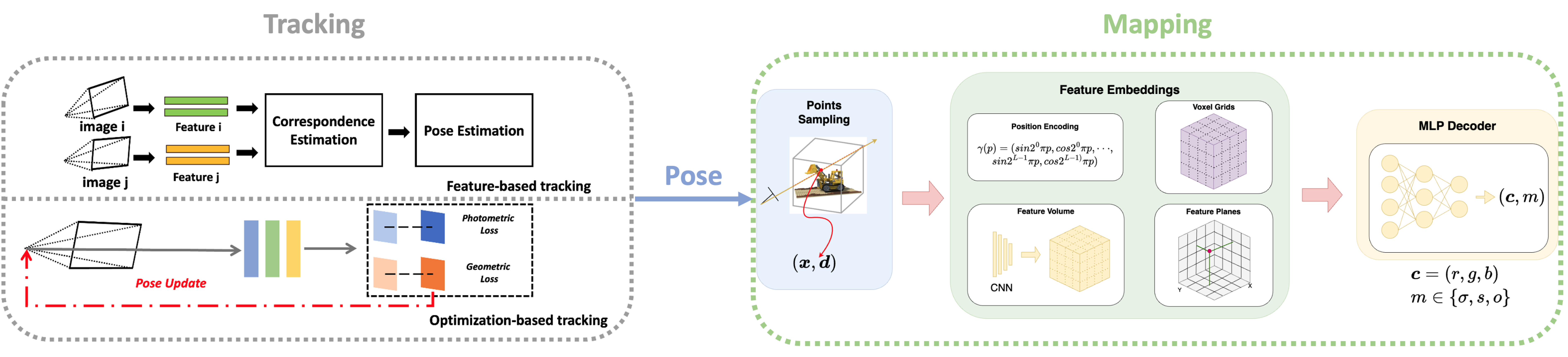}
    \caption{Overview of NeRF-based SLAM pipeline. Figure taken from ~\cite{ming2024benchmarking}.}
    \label{fig:nerf-slam}
\end{figure*}

They categorize NeRF-based methods in the SLAM domain into monocular SLAM, RGB-D SLAM, LiDAR SLAM, and multi-sensor SLAM. Additionally, they provide an overview of the NeRF-based SLAM pipeline, illustrating the framework and summarizing the various approaches used in NeRF-based SLAM methods which is detailed in Figure~\ref{fig:nerf-slam}. Figure~\ref{fig:nerf-slam} directly summarizes the current methods of applying NeRF in SLAM. 
% All NeRF-based SLAM approaches can be described using Figure~\ref{fig:nerf-slam}. 
Finally, they benchmark NeRF in the SLAM domain and provide comprehensive results to evaluate NeRF-based SLAM methods.

In contract to NeRF-based SLAM, 3D Gaussians~\cite{kerbl3Dgaussians} recently have emerged as a promising 3D scene representation. 3D Gaussian Splatting (3DGS) leverages the sparse nature of 3D environments by iterating over primitives for rasterization instead of tracing rays, effectively capturing the details of these scenes and speeding up rendering. There have been many recent advancements and applications of 3DGS, such as in the capture of static scenes where 3DGS introduces an innovative differentiable renderer known as Fuzzy Metaballs ~\cite{keselman2022approximate}. These facilitate low-latency differential rendering on CPUs and focus on depicting shapes through depth maps and silhouettes. In various vision-related tasks, particularly dynamic scene capture ~\cite{wu20244d}, 3DGS constructs an efficient Gaussian splatting framework that accommodates changes in Gaussian motion and shape over time, significantly improving dynamic scene capture. In the realm of 3D generation ~\cite{tang2023dreamgaussian}, it also adapts 3D Gaussian splatting into generative settings for 3D content creation, significantly reducing the generation time of optimization-based 2D lifting methods

In this paper, we aim to leverage the fast rendering performance and promising explicit 3D scene representation of 3DGS~\cite{kerbl3Dgaussians}, along with the accurate pose estimation strategy from Stereo-NEC~\cite{wang2024stereo}, to generate dense 3D reconstructions while maintaining accurate camera pose estimation~\cite{song2024eta}. We implement our pipeline using SplaTAM~\cite{spatam} as our foundation. 
The key contributions of our proposed method are as follows:
\begin{itemize}
\item We employ a rotation-translation decoupled strategy combined with inverse optimization to estimate accurate camera poses.
\item We utilize 3D Gaussian primitives to represent scenes, facilitating dense reconstruction.
\item  We utilize 3D Gaussian Splatting with rasterization, allowing for real-time differentiable optimization via CUDA.
\item  We incorporate depth priors as additional regularization to enforce geometric constraints, thereby enhancing the accuracy of both pose estimation and 3D reconstruction.
\end{itemize}

\section{PRELIMINARIES}
\subsection{Visual Geometry Residual}
The visual Geometry residual, \( r\mathcal{L}_{reproj} \) for camera frame at time $k$ is defined as:
\begin{equation*}
r\mathcal{L}_{reproj} = \textbf{x}^{i}- \pi_{(\cdot)}(\textbf{R}_{\text{c}_{\text{k}},\text{w}}\textbf{X}^{i}+\textbf{t}_{\text{c}_{\text{k}},\text{w}})
\end{equation*}
\begin{align*}\small
\pi_m\left(
\left[\begin{matrix} 
X \\
Y \\
Z
\end{matrix}\right]\right) = \left[\begin{matrix} 
f_x\frac{X}{Z} + c_x \\
f_y\frac{Y}{Z} + c_y
\end{matrix}\right],
\pi_s\left(
\left[\begin{matrix} 
X \\
Y \\
Z
\end{matrix}\right]\right) = \left[\begin{matrix} 
f_x\frac{X}{Z} + c_x \\
f_y\frac{Y}{Z} + c_y \\
f_x\frac{X-b}{Z} + c_x
\end{matrix}\right], 
\end{align*}
\(\textbf{X}^{i}\) is a 3D landmark in the world frame, derived through stereo matching using triangulation, while \(\textbf{x}^{i}\) indicates the corresponding 2D feature in image coordinate. The reprojection functions, \(\pi_{(\cdot)}\), include \(\pi_m\) for monocular reprojection and \(\pi_s\) for rectified stereo reprojection.  A 3D landmark is denoted as \([X, Y, Z]^\top\). Camera intrinsic parameter: \(f_x\) and \(f_y\) are the focal lengths, \(c_x\) and \(c_y\) denote the principal point, and \(b\) is the baseline of the stereo camera.
\subsection{Neural Radiance Fields}
NeRF's core concept revolves around modeling a 3D scene through neural networks, where the scene is encoded within the network' parameters. In practice, it utilizes multi-layer perceptrons (MLPs) to map spatial locations to an implicit representation of the surfaces of target objects or scenes. Moreover, the vanilla NeRF approach places a strong emphasis on volume density.
\begin{figure}[!hb]
    \centering
    \includegraphics[width=\linewidth]{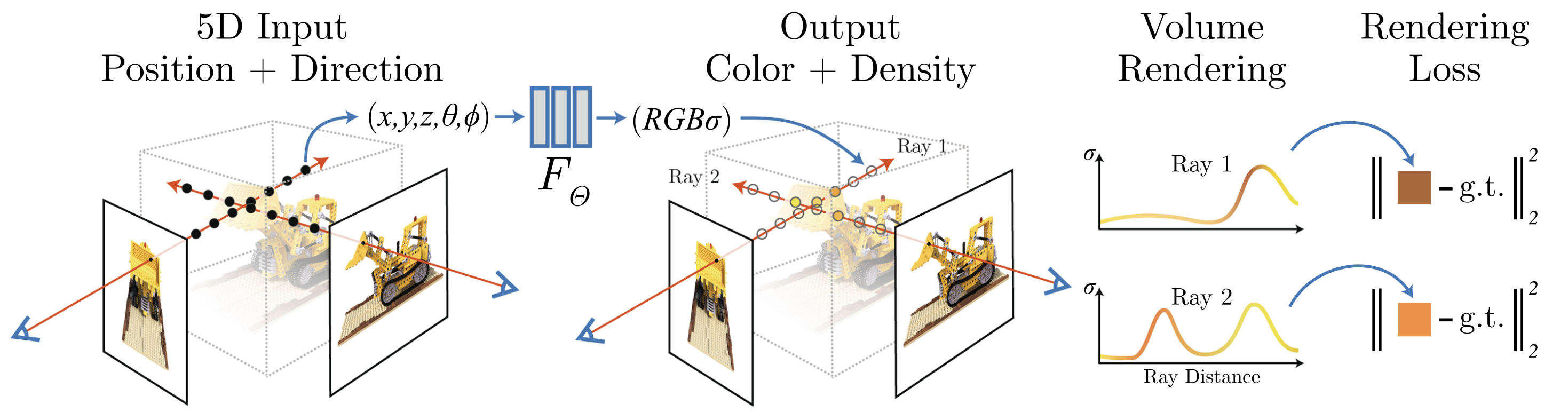}
    \caption{Pipeline of NeRF. Figure taken from \cite{nerf-c}.}
    \label{fig:nerf-pipeline}
\vspace{-4ex}
\end{figure}
\subsection{3D Gaussians Splatting}
3D Gaussian Splatting is another technique that allows for novel view synthesis and 3D reconstruction. Unlike NeRF, which uses implicit representation for the environment, 3D Gaussian Splatting employs 3D Gaussians as primitives for explicit 3D scene representation. This process allows for fast, differentiable rendering of 3D scenes by projecting the 3D Gaussians onto a pixel-based image plane, resulting in a rendered image.
Each Gaussian is represented by Equation~\ref{gaussian} and is parameterized by only eight values: three for its RGB color $c$, three for its center position $\boldsymbol{\mu} \in \mathbb{R}^3$, one for its radius $r$, and one for its opacity $o \in [0, 1]$. Since 3D Gaussian splatting is differentiable, it allows for the updating of the eight parameters using gradient-based optimization by minimizing the difference between the rendered image and the input image. 
\begin{equation}\label{gaussian}
f(\textbf{p}) = \textit{o}\exp(-\frac{||\textbf{p}-\boldsymbol\mu||^2}{2r^2})
\end{equation}
\textbf{Render an RGB image}. 
Gaussian splatting renders an RGB image by alpha-compositing as follows:
\begin{equation}
\textit{C}(\textbf{p}) = \sum_{i=1}^{\text{n}}c_i f_i(\textbf{p})\prod_{j=1}^{i-1}(1-f_j(\textbf{p}))
\end{equation}
here, $f_i(\cdot)$ refers to the function defined in Equation~\ref{gaussian}, and $c_i$ represents the color of the \(i\)-th 3D Gaussian.
\section{Proposed Approach}\label{sec:vi_3dgs_method}
Our method inspired by Stereo-NEC~\cite{wang2024stereo}, SplaTAM~\cite{spatam}, and BundledSLAM~\cite{10503743} aims to provide a SLAM system with dense reconstruction using depth priors. During the tracking stage, we adopt the decoupled rotation and translation estimation strategy from Stereo-NEC, using inverse optimization. Additionally, we incorporate the 3DGS concept from SplaTAM for 3D scene representation, which enables the ability to synthesize novel views. Multi-camera SLAM systems have many advantages, quite directly resulting from the ability to incorporate information coming from a larger field of view—leading to higher robustness and better localization accuracy. This work presents a significant extension and enhancement of the state-of-the-art stereo SLAM system, ORB-SLAM2, with the goal of achieving even greater precision. To achieve this, first, measurements from all cameras are projected to a common virtual camera, BundledFrame. The virtual camera is properly designed to facilitate hassle-free handling of multi-camera setups and an opportunity to properly combine data obtained from individual cameras. The benchmarks include an extensive comparison with ground truth measurements made on the state of the art SLAM system. The assessment is performed on the EuRoC datasets to evaluate the performance of the system critically. The results demonstrate the superior accuracy of the BundledSLAM system compared to existing approaches. Specifically, building on Stereo-NEC, SplaTAM, and BundledSLAM, we separately estimate rotation and translation using inverse optimization. This approach directly employs differentiable rendering and gradient-based optimization to refine both the camera pose and the parameters of the 3D Gaussian map, based on the set of online camera poses estimated so far.

Our method involves four steps designed to derive precise initial values for camera state estimation and 3D Gaussian parameter optimization. These steps are crucial for scene representation and encoding the environment:
\begin{itemize}
    \item \textbf{Step 1. Initialization}: We initialize initial 3D Gaussian parameter by using first camera frame. 
    \item \textbf{Step 2. Pose Estimation}: This step aims to estimate camera pose incrementally.
    \item \textbf{Step 3. Scene Optimization}: This step aims to update the parameters of the 3D Gaussian map based on the set of online camera poses estimated so far.
\end{itemize}

\subsection{Initialization}\label{sec:step1}
For the first frame, we set the camera pose as an identity matrix and initialize new Gaussians for all pixels. For each pixel, we create a new Gaussian with the corresponding color of that pixel. The center of each Gaussian is set at the location obtained by unprojecting the pixel using its corresponding depth from the depth map. Each Gaussian is assigned an opacity of 0.5 and a radius such that it appears as a one-pixel radius in the 2D image, which is calculated by dividing the depth by the focal length.

\subsection{Pose Estimation}\label{sec:step2}
Pose estimation occurs during the tracking stage, aiming to estimate the camera pose for each frame in real time. To enable accurate pose estimation and real-time performance, we use a different approach from the original SplaTAM, which assumes constant velocity for the initial guess and then updates the current camera pose via inverse optimization. This method can provide inaccurate pose estimates during large motions or intense rotations, leading to inferior pose estimation later.

In response, we utilize a decoupled rotation and translation strategy, iteratively updating both in several iterations through gradient-based optimization. This process involves differentiably rendering RGB, depth, and silhouette maps and updating the camera parameters to minimize a combined loss of photometric loss, depth geometry loss, and visibility loss. The rendered depth and rendered silhouette for visibility are illustrated in Equations~\ref{eq_dr} and ~\ref{eq_sr}, respectively. This method helps maintain tracking performance, and with 3D Gaussian Splatting, it also supports real-time performance during gradient-based optimization through differentiable rendering.

\textbf{Render a depth image.}
Gaussian splatting renders a Depth image as follows:
\begin{equation}\label{eq_dr}
\textit{D}(\textbf{p}) = \sum_{i=1}^{\text{n}}d_i f_i(\textbf{p})\prod_{k=1}^{i-1}(1-f_k(\textbf{p}))
\end{equation}
where $d_i$ denotes the depth of the \(i\)-th 3D Gaussian in camera frame.

\textbf{Render a silhouette image.}
Gaussian splatting renders a silhouette image, which is used to determine visibility, as follows:
\begin{equation}\label{eq_sr}
\textit{S}(\textbf{p}) = \sum_{i=1}^{\text{n}} f_i(\textbf{p})\prod_{k=1}^{i-1}(1-f_k(\textbf{p}))
\end{equation}

\begin{table*}[t] %[!tbp]
\centering
{
\vspace{0.20in}
\caption{%\footnotesize 
Quantitative results for evaluating localization ($\downarrow$), novel view synthesis ability ($\uparrow$), and geometric accuracy ($\downarrow$).
}
\label{tab:qauntitative_res}
 \resizebox{\textwidth}{!}{
\begin{tabular}{ccccccc}
    \toprule
    \multicolumn{1}{c}{ Sequence} & \multicolumn{1}{c}{ Final Average ATE RMSE ($\downarrow$)} & \multicolumn{1}{c}{ Average PSNR ($\uparrow$)} & \multicolumn{1}{c}{ Average Depth RMSE ($\downarrow$)}  \cr
    
    \midrule
    freiburg1\_desk & 3.49 cm & 21.20 & 3.61 cm\cr
    freiburg1\_desk2 & 6.58 cm & 20.74 & 3.24 cm\cr
    freiburg1\_room & 12.52 cm & 19.63 & 3.54 cm\cr
    freiburg2\_xyz & 1.41 cm & 25.11 & 2.81 cm\cr
    freiburg3\_long\_office\_household & 7.92 cm & 20.99 & 6.00 cm\cr
    \bottomrule
    \end{tabular}
}
}
\end{table*}

\begin{table}[t] %[!tbp]
\centering
{
\vspace{0.20in}
\caption{%\footnotesize 
Quantitative results for evaluating depth ($\downarrow$), Structural Similarity Index Measure (SSIM $\uparrow$), and Learned Perceptual Image Patch Similarity (LPIPS $\downarrow$).
}
\label{tab:qauntitative_res}
 \resizebox{\columnwidth}{!}{
\begin{tabular}{ccccccc}
    \toprule
    \multicolumn{1}{c}{ Sequence} & \multicolumn{1}{c}{ Average Depth Loss} & \multicolumn{1}{c}{ Average MS-SSIM} & \multicolumn{1}{c}{ Average LPIPS}  \cr
    
    \midrule
    freiburg1\_desk & 3.61 cm & 0.846 & 0.244\cr
    freiburg1\_desk2 & 3.24 cm & 0.839 & 0.253\cr
    freiburg1\_room & 3.54 cm & 0.810 & 0.252\cr
    freiburg2\_xyz & 2.81 cm & 0.955 & 0.073\cr
    freiburg3\_long\_office\_household & 6.00 cm & 0.857 & 0.204\cr
    \bottomrule
    \end{tabular}
}
}
\end{table}

\subsection{Scene Optimization}\label{sec:step3}
In this section, we aim to reconstruct a dense  representation by a set of explicit 3D Gaussians and then update the scene through 3D Gaussian Splatting. As camera poses are updated and new Gaussians are added, there may be duplicate 3D Gaussians that need to be removed. Additionally, optimizing the parameters of the 3D Gaussians is essential for accurate scene representation. Therefore, our scene optimization process includes both 3D Gaussian parameter optimization and duplication removal.

Specifically, the inverse optimization in this step is performed using differentiable rendering and gradient-based optimization, similar to the process in the tracking stage~\ref{sec:step2}. However, unlike pose estimation, the camera poses remain fixed, and the parameters of the Gaussians are updated. To ensure real-time performance, we do not use all frames for scene optimization. Instead, we use only keyframes to update the 3D Gaussian parameters and to handle duplication. This means new Gaussians are added only when the current frame is a keyframe.

Our keyframe selection strategy is based on the average parallax from the previous keyframe. If the average parallax of tracked features between the current frame and the latest keyframe exceeds a certain threshold, the current frame is designated as a new keyframe.
\section{EVALUATION}
\subsection{Experiment Setup}
Our proposed method is benchmarked using the TUM-RGBD dataset~\cite{6385773}. TUM-RGBD is a well-known benchmark for evaluating RGB-D SLAM systems. It includes a large set of image sequences recorded with a Microsoft Kinect, along with highly accurate and time-synchronized ground truth camera poses provided by a motion capture system. The sequences contain both color and depth images at a full sensor resolution of 640 × 480, captured at a video frame rate of 30 Hz. The ground truth trajectory is obtained using a motion capture system equipped with eight high-speed tracking cameras (100 Hz).

The dataset consists of 39 sequences recorded in various environments, including an office and an industrial hall, covering a wide range of scenes and camera motions. TUM-RGBD is a challenging dataset for RGB-D SLAM, especially for dense methods, due to the poor quality of both the RGB and depth images, as they are captured using older, lower-quality cameras. We chose this dataset to evaluate our method due to its difficulty.

Our method is implemented on a system with an Intel i7-9700k @ 3.6 GHz processor, 64 GB of RAM, and an Nvidia GeForce RTX 4080 SUPER GPU with 16 GB of memory.
\subsection{Experiment Results}
\begin{figure}[h]
    \centering
    \includegraphics[width=\linewidth]{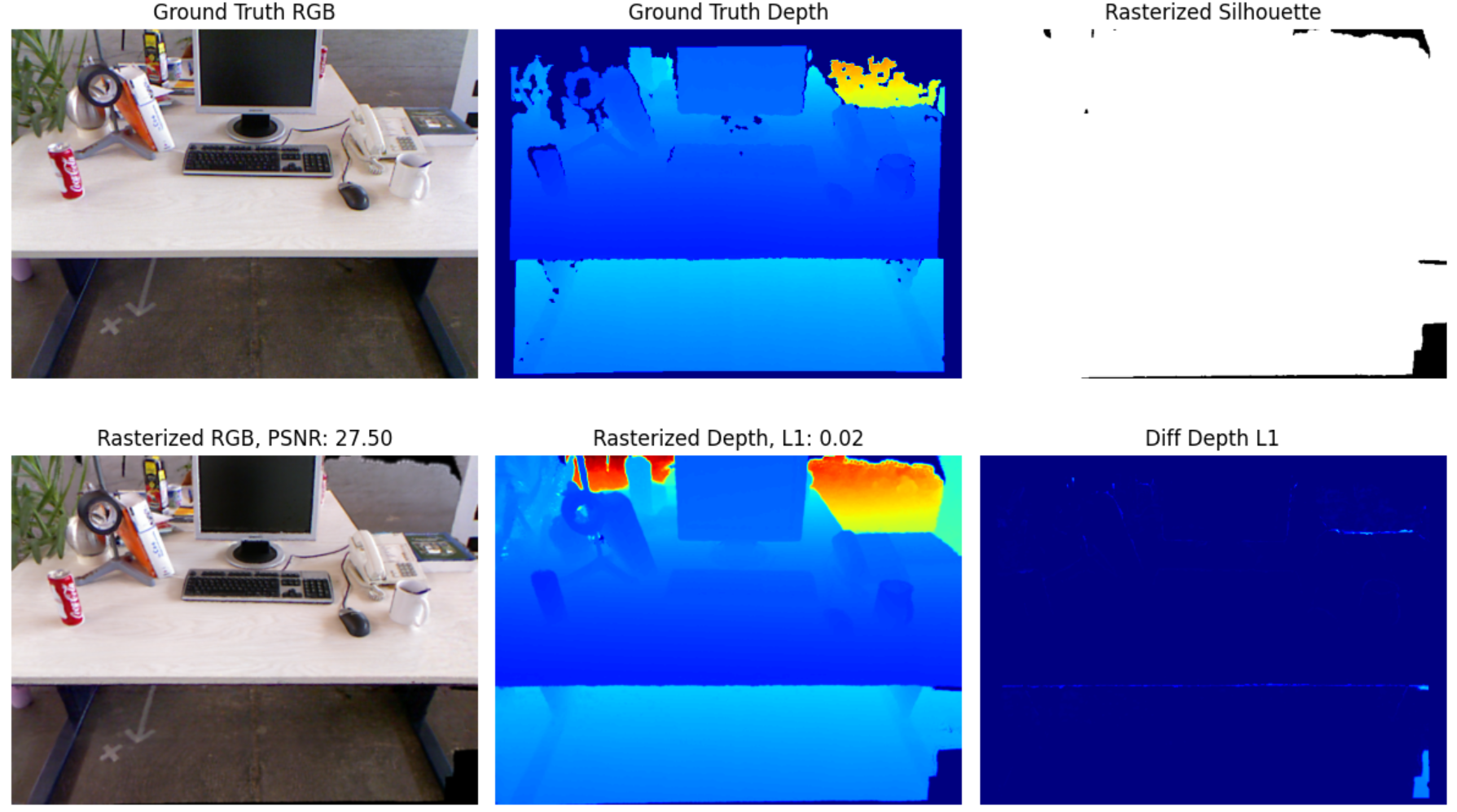}
    \caption{
Qualitative rendering results on freigburg2. The top row, from left to right, shows the input RGB, input depth map, and rasterized silhouette. The bottom row, from left to right, displays the rendered RGB, rendered depth map, and visualization of the L1 loss on the depth map.}
    \label{fig:fig1}
% \vspace{-4ex}
\end{figure}
\textbf{Quantitative Evaluation:}
We evaluate our method on five different challenging scenarios from the TUM-RGBD dataset. We assess the localization performance of our method using the absolute trajectory error (ATE), the novel view synthesis performance using the peak signal-to-noise ratio (PSNR), and the geometric representation accuracy using the root mean square error (RMSE) of the depth map. 
\begin{figure}[h]
    \centering
    \includegraphics[width=\linewidth]{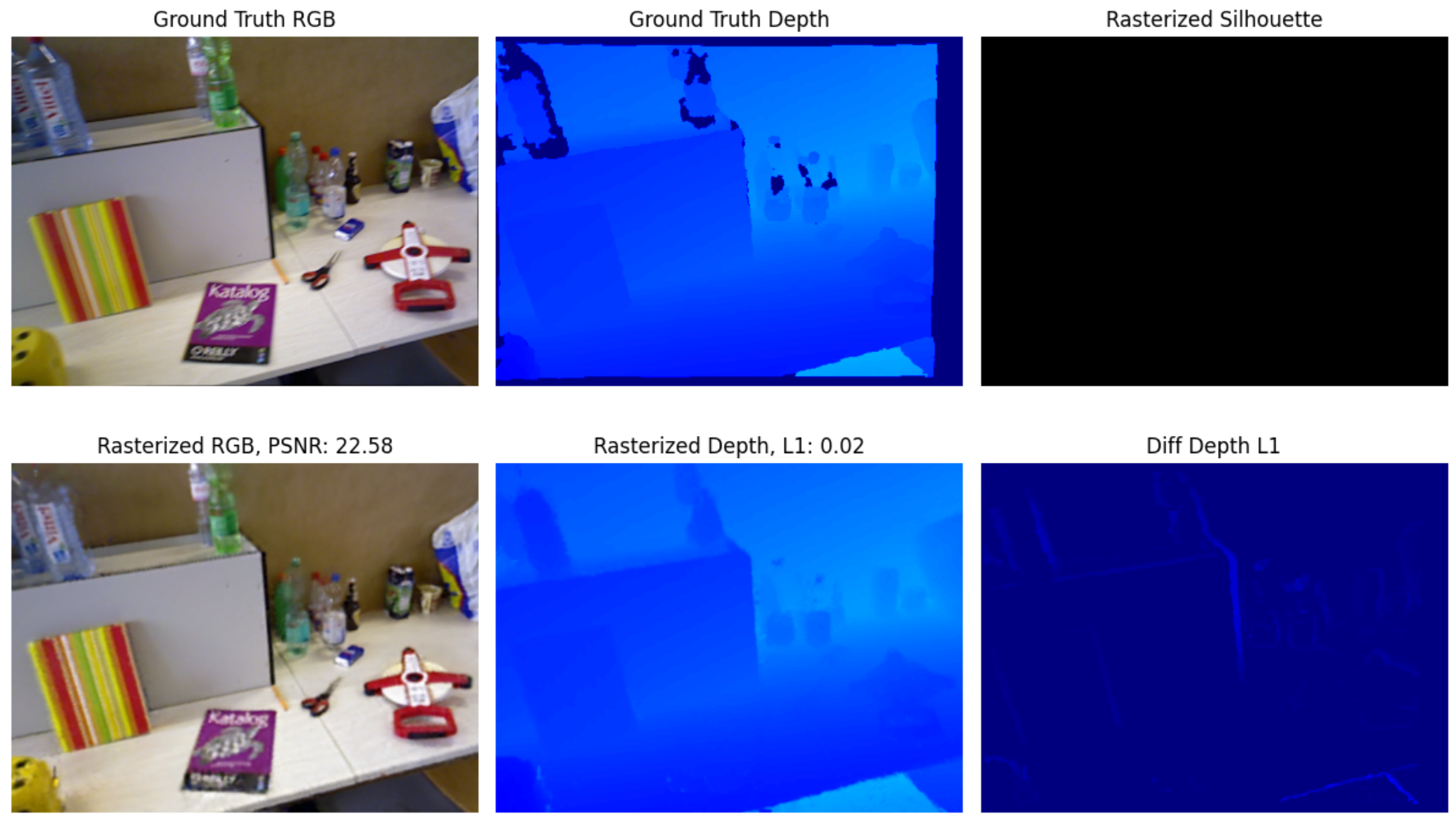}
    \caption{Qualitative rendering results on freiburg3 office. The top row, from left to right, shows the input RGB, input depth map, and rasterized silhouette. The bottom row, from left to right, displays the rendered RGB, rendered depth map, and visualization of the L1 loss on the depth map.}
    \label{fig:fig2}
% \vspace{-4ex}
\end{figure}

\begin{figure}[h]
    \centering
    \includegraphics[width=\linewidth]{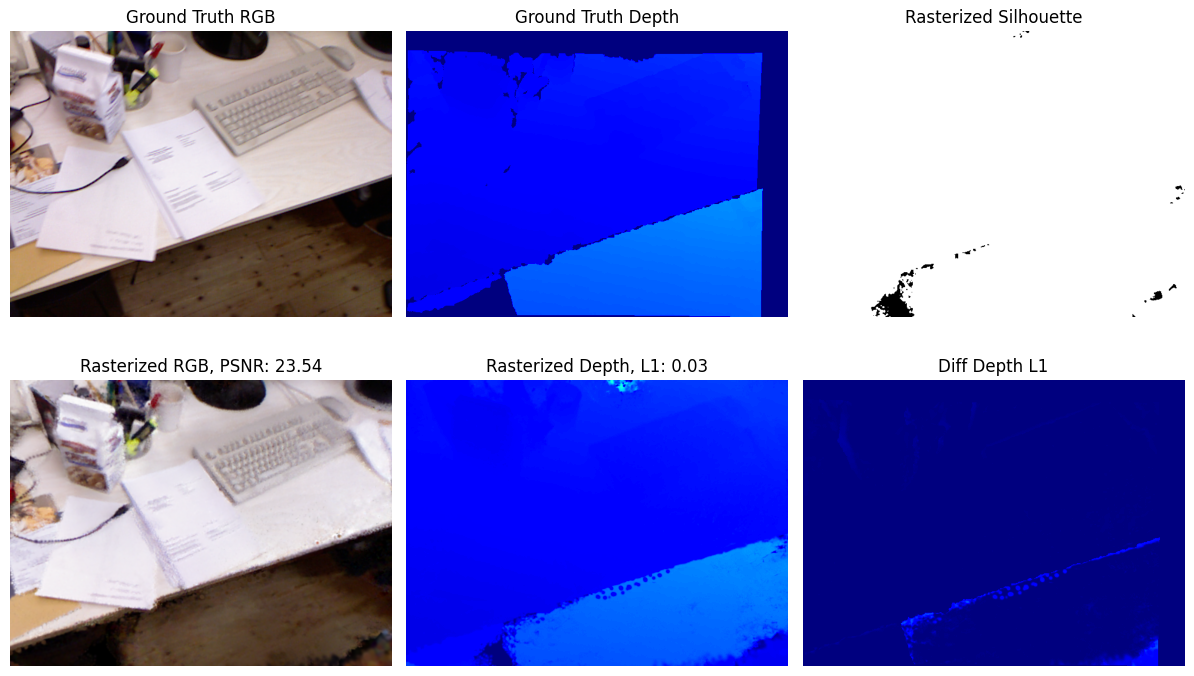}
    \caption{Qualitative rendering results on freiburg1 desk. The top row, from left to right, shows the input RGB, input depth map, and rasterized silhouette. The bottom row, from left to right, displays the rendered RGB, rendered depth map, and visualization of the L1 loss on the depth map.}
    \label{fig:fig2}
% \vspace{-4ex}
\end{figure}

\begin{figure}[h]
    \centering
    \includegraphics[width=\linewidth]{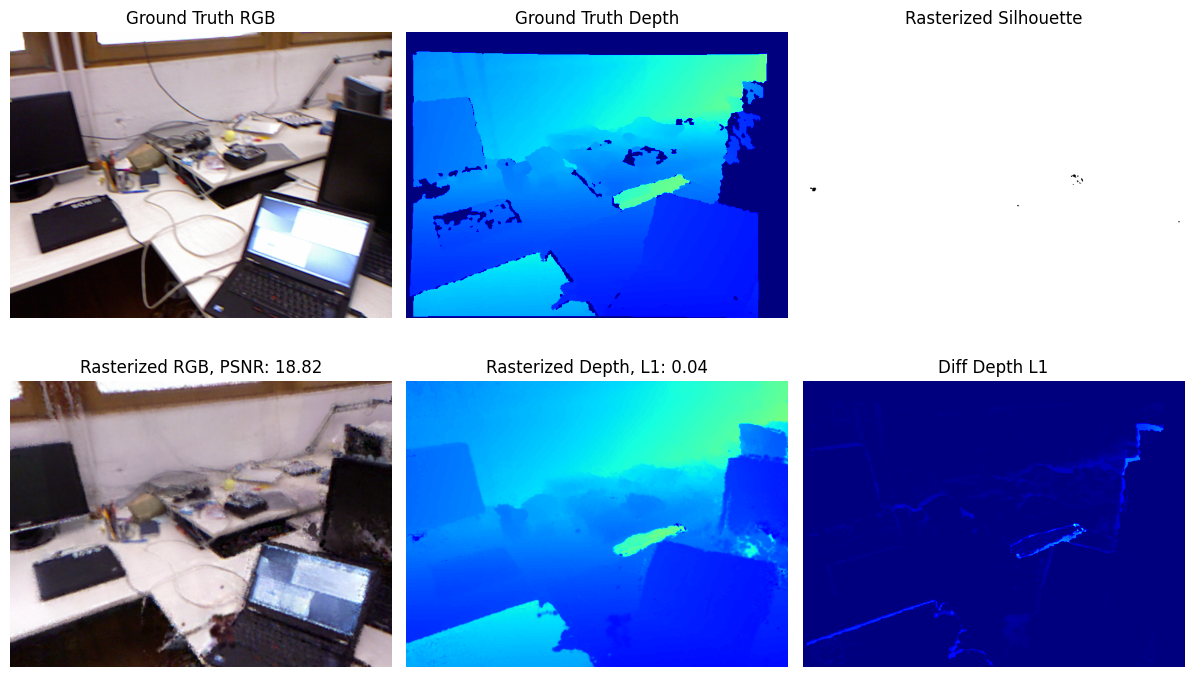}
    \caption{Qualitative rendering results on freiburg1 room. The top row, from left to right, shows the input RGB, input depth map, and rasterized silhouette. The bottom row, from left to right, displays the rendered RGB, rendered depth map, and visualization of the L1 loss on the depth map.}
    \label{fig:fig2}
% \vspace{-4ex}
\end{figure}

Our method achieved localization accuracy with an average ATE ranging from 3.49 cm to 12.52 cm across five challenging scenarios, demonstrating its capability to deliver centimeter-level precision in difficult indoor environments. Absolute trajectory error (ATE) is a widely used metric for assessing localization accuracy when benchmarking the tracking performance of SLAM systems. The superior accuracy of our approach is evident in the average ATE values presented in Table~\ref{tab:qauntitative_res}.

We also evaluate the novel view synthesis performance of our method using PSNR, with values ranging from 19.63 to 25.11, indicating competitive performance. Additionally, we assess geometric accuracy by computing the average depth map RMSE, which showed depth accuracy ranging from 2.81 cm to 6 cm. These strong performances in PSNR and depth map accuracy are highlighted in the average PSNR and average Depth RMSE columns of Table~\ref{tab:qauntitative_res}, respectively.

\textbf{Qualitative Evaluation:}
Alongside the quantitative evaluation, we offer a qualitative assessment as well. As illustrated in Fig~\ref{fig:fig1}, we compare our rendered RGB and depth images on the Freiburg2\_xyz sequence with the input images shown in the top row. Additionally, we visualize the silhouette map to highlight areas of new observations in the current frame where new 3D Gaussians will be added to the map. Specifically, the white regions indicate where these new 3D Gaussians will be incorporated.

Similarly, Fig~\ref{fig:fig2} presents our rendered results for both RGB and depth images on the Freiburg3\_long\_office sequence, compared with the input images in the top row. The silhouette map in Fig~\ref{fig:fig2} shows that there are no new observations from the current frame's viewpoint on the Freiburg3\_long\_office sequence.

\section{CONCLUSIONS}
In this paper, we introduce an innovative RGB-D SLAM system that utilizes 3D Gaussian Splatting (3DGS) for accurate point cloud generation, enabling dense reconstruction and real-time pose estimation. Our system takes advantage of the computational efficiency of CUDA-enabled rasterization to render both RGB and depth images. It optimizes camera poses and 3D Gaussians by minimizing the discrepancy between the rendered images and the input images. Additionally, through a rotation-translation decoupled strategy, our method successfully meets the demanding requirements of real-time performance and precise scene reconstruction.

We also provide extensive evaluations on the challenging TUM-RGBD dataset demonstrate that our method not only meets but exceeds existing benchmarks in terms of pose accuracy, geometric accuracy, and rendering performance. The proposed approach shows significant promise, achieving centimeter-level localization precision and enhanced depth accuracy in diverse indoor settings. This underlines the robustness and reliability of our system even in environments with poor image quality, a common challenge in practical applications.

Furthermore, the integration of depth priors as regularization effectively mitigates the inherent limitations of 3D Gaussian representations, such as multi-view inconsistencies, thereby enhancing both the accuracy of pose estimations and the fidelity of 3D reconstructions. Our results demonstrate the robustness of the proposed method and its potential to significantly advance the field of visual SLAM.

Future work could explore the scalability of our approach in larger and more dynamic environments, with the possibility of incorporating advanced machine learning algorithms to further refine the 3DGS technique and broaden its application. The promising results from our study encourage ongoing research towards fully autonomous systems capable of operating in real-world scenarios with the same level of precision and efficiency observed in controlled experimental settings. This work represents a notable advancement in the integration of dense 3D reconstruction within real-time SLAM systems, opening new avenues for immersive and interactive technologies in robotics, augmented reality, and beyond.

\printbibliography
\end{document}